# Training Semantic Descriptors for Image-Based Localization

Ibrahim Cinaroglu[1] and Yalin Bastanlar[1]

[1]Computer Engineering Dept., Izmir Institute of Technology, Urla, Izmir, Turkey
ibrahim.ceinaroglu@gmail.com,   yalinbastanlar@iyte.edu.tr

**Abstract.** Vision based solutions for the localization of vehicles have become popular recently. We employ an image retrieval based visual localization approach. The database images are kept with GPS coordinates and the location of the retrieved database image serves as an approximate position of the query image. We show that localization can be performed via descriptors solely extracted from semantically segmented images. It is reliable especially when the environment is subjected to severe illumination and seasonal changes. Our experiments reveal that the localization performance of a semantic descriptor can increase up to the level of state-of-the-art RGB image based methods.

## 1   Introduction

Visual localization can be defined as estimating the position and orientation of a visual query material within a known environment. Locating an agent (could be a pedestrian or a vehicle) is critical for city-scale navigation and other location-based services. In our work, we propose a method for localization based on image retrieval. Our method utilizes a geotagged database images which are kept with GPS coordinates and the known geographic location of the retrieved database image (best match) serves as an approximate position of the query image. The novelty of our work is that we perform image retrieval on semantically segmented images rather than RGB images in order to increase localization performance. To do that we train a semantic descriptor. Some recent studies showed that semantic cues can be used to improve localization accuracy but no previous study directly performed localization using semantic labels of the database images.

## 2   Previous Work and Our Contributions

Classical methods of image retrieval based localization mostly depend on Bag-of-Features [5], where SIFT-like features extracted from all images in the database are clustered to define a set of 'visual words', then an image is represented by a visual word frequency histogram. Recent studies, however, use features from the convolutional layers of convolutional neural networks (CNNs) [3]. In NetVLAD [1], a trainable CNN, a specially designed layer is added to a standard CNN to convert the last convolutional layer into a compact descriptor. In their study, NetVLAD outperformed state-of-the-art techniques based on experiments performed on four different datasets.

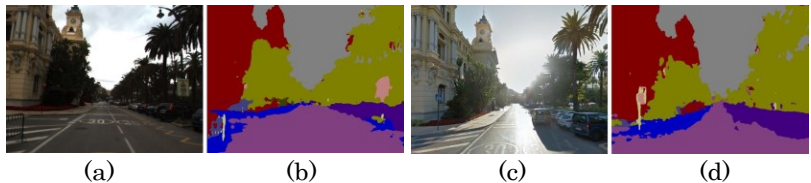

(a)              (b)              (c)              (d)

**Fig. 1.** (a) and (c) are two images of the same scene with drastic illumination changes. (b) and (d) are their semantic segmentations, respectively. Standard methods have low performance for such cases, where more stable semantic segmentation can help.

Our work is based on the hypothesis that semantic decomposition of a scene can increase localization performance. We can rely on the semantic labels especially when there are illumination and long-term changes in the scene. As seen in Fig. 1, weather conditions (sunny, cloudy etc.) or time of the day can result in drastic appearance changes. RGB image based retrieval methods face difficulties in such cases, whereas semantic segmentation can give more stable results.

The idea of using semantic labels to improve image based localization has been partially explored. In [9], point matches are checked if their semantic labels are also matching. In [4], features are extracted with a CNN, but a weighting scheme is applied based on semantic labels (e.g. increasing weights for buildings since they are more stable in long time intervals). An attempt to design a descriptor from 2D semantic labels was first proposed in [7] but rather than localization the descriptor was used to distinguish street intersections from other scenes. To the best of our knowledge, our study is the first to perform localization with semantically segmented images.

Stenborg et al. [8] performed localization based on the query image's semantic content when the environment is 3D reconstructed and semantically labeled. This is an innovative study in terms of performing localization purely based on semantic labels; however it requires semantic labels of 3D point clouds. which is not available in most cases. Our method employs 2D images and it is much cheaper than the approaches requiring semantic 3D reconstruction [8, 6].

## 3   Our Method

We aim to perform localization with the descriptors extracted from semantically segmented images. In our study, semantic segmentation results of database images are given to a CNN as the training set to minimize a triplet loss function (Fig. 2). In triplet loss, for a given input image, images taken from a similar location constitute a positive set and images from far away positions constitute a negative set. Thus, a descriptor is optimized so that the distance to the positive set is minimized and the distance to the negative set is maximized.

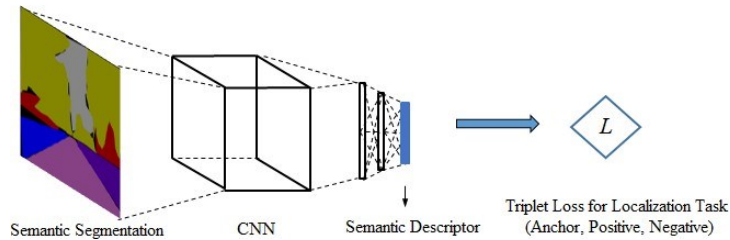

**Fig. 2.** Semantic segmentation images are given to CNN to train a semantic descriptor

## 4 Experiments

We performed experiments on RobotCar Seasons Dataset[1], a benchmark dataset with various appearance changes (such as sunny/cloudy or summer/winter). Dataset was collected in Oxford, UK by passing through the same route over a period of 12 months. We used rear-view reference images as the database (2318 images) and overcast-winter images as the query set (390 images).

Semantic labels in our approach are extracted with DeepLabv3+ [2] retrained on CamVid and weakly-supervised segmented RobotCar Seasons dataset. For the proposed approach of training semantic descriptor (Fig. 2), we used AlexNet as the backbone architecture.

Performance of the proposed approach is compared with a state-of-the-art RGB image based method (NetVLAD [1]). Performance is measured with two metrics:

- Top-1 Recall @D: Distance(D) between the top ranked database image position and the query ground truth position is calculated.

- Recall @N: Percentage of well localized (≤25m distance error) queries within the top N number of returned candidates.

Results are given in Fig. 3. RGB image based method seems to be slightly better than the semantic based method but according to Recall @N metric, semantic method is better after N=4. A localization example where RGB image based method fails but semantic based approach accomplishes is given in Fig. 4. One can observe the challenging illumination conditions.

## 5 Conclusions and Future Work

We trained a semantic segmentation based descriptor to be used for image based localization. Its performance came closer to the state-of-the art RGB image based localization methods. This study serves as a proof that the semantic labels carry a great potential for localization. In near future, we plan to merge RGB image based method and semantic method in a hybrid fashion with the expectation of results that are better than using solely one approach.

---

[1] https://data.ciirc.cvut.cz/public/projects/2020VisualLocalization/RobotCar-Seasons/

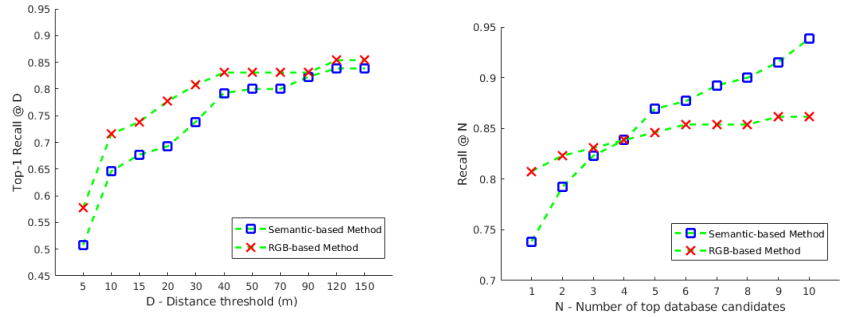

**Fig. 3.** Top-1 Recall @D results (left) and Recall @N results (right) with RobotCar Seasons Dataset (overcast-winter query set)

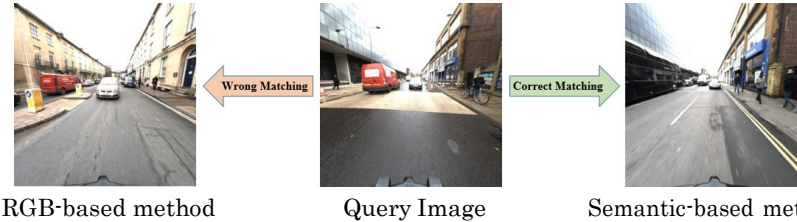

RGB-based method          Query Image          Semantic-based method

**Fig. 4.** An example query (middle) that RGB image method (left) fails but semantic descriptor (right) does not fail